\documentclass[11pt,a4paper]{article}

\usepackage[T1]{fontenc}
\usepackage[utf8]{inputenc}
\usepackage{lmodern}
\usepackage[margin=1in]{geometry}
\usepackage{microtype}
\usepackage{placeins}
\usepackage{graphicx}
\graphicspath{{./}}
\usepackage{amsmath,amssymb,amsfonts,mathtools,amsthm}
\usepackage{booktabs,multirow,array,tabularx}
\usepackage{textcomp}
\usepackage[table]{xcolor}
\usepackage{pifont}
\usepackage{siunitx}
\usepackage[numbers,sort&compress]{natbib}
\usepackage[hidelinks]{hyperref}
\usepackage[capitalize,noabbrev]{cleveref}
\hypersetup{bookmarksdepth=3}

\definecolor{myteal}{RGB}{120,180,170}
\definecolor{dwarecip}{RGB}{255,236,210}
\definecolor{oursrecip}{RGB}{232,224,245}

\newcommand{\greencmark}{\textcolor{green!60!black}{\ding{51}}}
\newcommand{\redxmark}{\textcolor{red}{\ding{55}}}
\newcolumntype{Y}{S[table-format=3.2]}
\newcolumntype{Z}{S[table-format=3.2(2)]}

\theoremstyle{plain}

\theoremstyle{definition}

\theoremstyle{remark}

\title{CARE: Consistent Action--Response Execution for Human-Guided Robot Transport}
\newcommand{\careauthors}{%
\begin{minipage}{\dimexpr\textwidth-2\tabcolsep\relax}
\centering
Ning Liu\textsuperscript{1}, Sen Shen\textsuperscript{2}, Zheng Li\textsuperscript{1},
Matthew D'Souza\textsuperscript{3},\\
Jen Jen Chung\textsuperscript{3,*}, Lan Wu\textsuperscript{1}, Thomas Br{\"a}unl\textsuperscript{1}
\par\vspace{0.7em}
\small
\textsuperscript{1}School of Engineering, The University of Western Australia,\\
35 Stirling Highway, Perth, WA 6009, Australia
\par\vspace{0.4em}
\textsuperscript{2}Department of Mechanical and Automation Engineering,\\
The Chinese University of Hong Kong, Shatin, Hong Kong SAR, China
\par\vspace{0.4em}
\textsuperscript{3}School of Electrical Engineering and Computer Science,\\
The University of Queensland, Brisbane, QLD 4072, Australia
\par\vspace{0.7em}
\footnotesize
\href{mailto:ning.liu@research.uwa.edu.au}{\nolinkurl{ning.liu@research.uwa.edu.au}},
\href{mailto:senshen@link.cuhk.edu.hk}{\nolinkurl{senshen@link.cuhk.edu.hk}}\\
\href{mailto:zheng.li@research.uwa.edu.au}{\nolinkurl{zheng.li@research.uwa.edu.au}},
\href{mailto:m.dsouza@uq.edu.au}{\nolinkurl{m.dsouza@uq.edu.au}}\\
\href{mailto:jenjen.chung@uq.edu.au}{\nolinkurl{jenjen.chung@uq.edu.au}},
\href{mailto:lan.wu@uwa.edu.au}{\nolinkurl{lan.wu@uwa.edu.au}},
\href{mailto:thomas.braunl@uwa.edu.au}{\nolinkurl{thomas.braunl@uwa.edu.au}}
\par\vspace{0.4em}
\textsuperscript{*}Corresponding author: Jen Jen Chung
\end{minipage}%
}
\author{\careauthors}
\date{}

\begin{document}
\maketitle

\begin{abstract}
Human-guided collaborative transport becomes interaction-critical when the operator can continue along a route segment that is not executable by the robot--payload system. In this regime, operator motion is not an exogenous reference: it simultaneously communicates a persistent route tendency and adapts to the robot's realized progress. Yet a common modular formulation predicts one operator future from observation history, scores multiple robot proposals against that shared forecast, and only afterward modifies the selected command for local feasibility. This can compare candidates under responses they would not induce and can condition prediction on an action that is never executed. We formulate local control over executable-action-indexed joint human--robot transitions and instantiate it in Consistent Action--Response Execution (CARE). Each nominal proposal or bounded alternative is first mapped to an
executable action; that same action conditions the route-aligned
operator-response prediction, the matched robot--operator rollout, and
a fixed interaction utility used for candidate selection. Bounded reciprocal refinement handles transient route--feasibility mismatch, whereas persistent progress loss replaces the nominal proposal with a progress-oriented recovery proposal while retaining the same downstream executable-action-conditioned evaluation. Across training and held-out maps, CARE preserves near-saturated performance when the indicated route remains locally usable and achieves the highest observed success and service availability when the operator can continue but the robot must detour. It also yields lower jerk-proxy values and measured controller latency than dynamic-window and Timed Elastic Band baselines. Factorial ablations distinguish reciprocal refinement from progress recovery, delayed-response tests probe temporal mismatch, and an indoor deployment demonstrates physical executability.\end{abstract}


\section{Introduction}
\label{sec:introduction}

\begin{figure}[ht]
    \centering
    \includegraphics[width=\columnwidth]{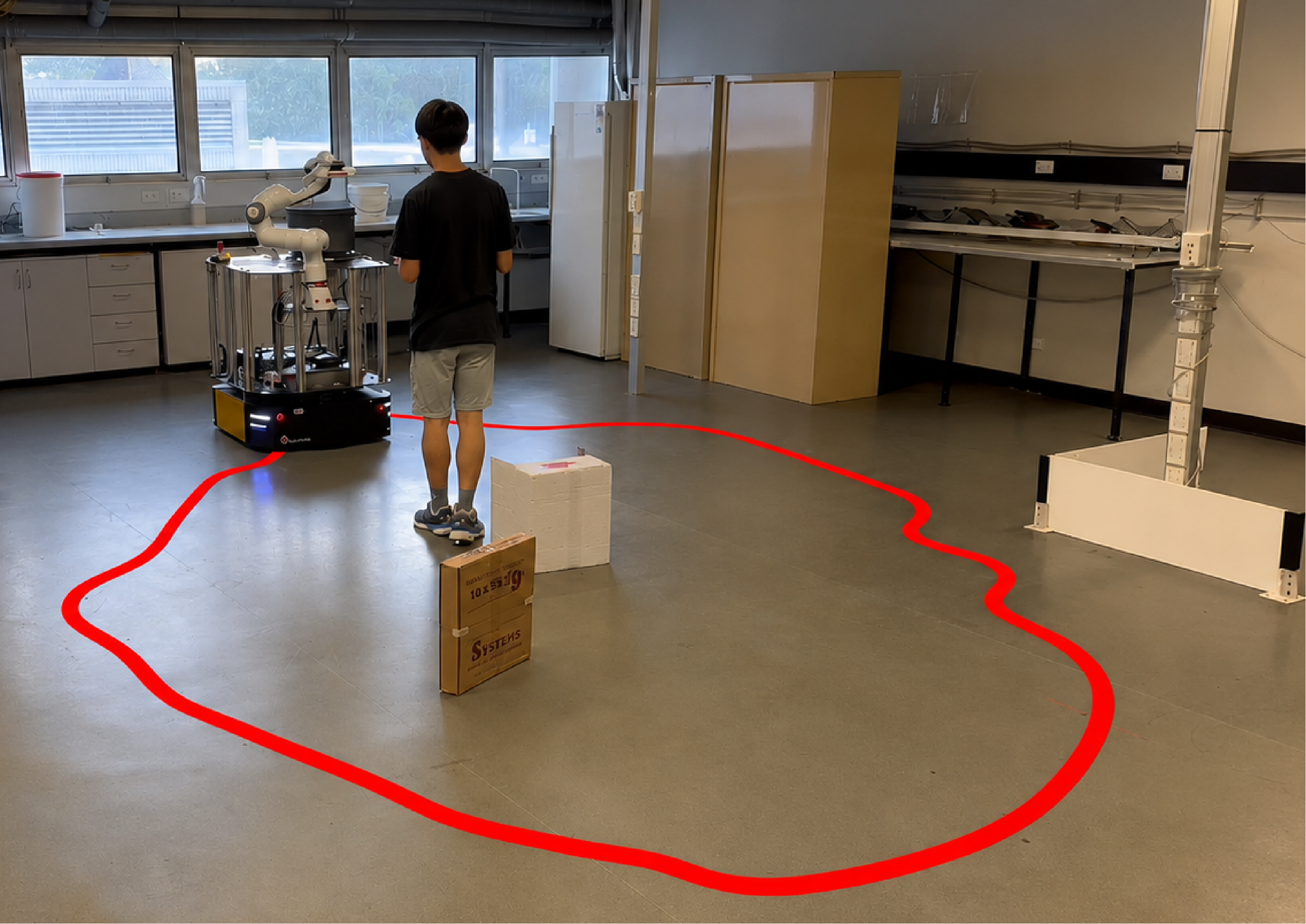}
    \caption{Physical human-guided transport with CARE. The red curve shows
    the executed robot trajectory while negotiating a local obstacle}
    \label{fig:realworld}
\end{figure}

Human-guided collaborative transport distributes navigation authority
asymmetrically between a human operator and a mobile robot. The operator
communicates task-level route guidance implicitly through motion, whereas the
robot carries or supports the payload and must realize that guidance under its
own geometry, motion, and local safety constraints. This asymmetric
leading--following relationship is closely related to human-following and
human-guidance mobile robotics, where human motion provides navigation and
interaction cues \cite{eirale2025human}. More broadly, preferred
leader--follower roles can shape perceived human--robot collaboration
\cite{noormohammadi2025human}. Existing systems have demonstrated onboard
person following and shared-control operation in unknown environments without
requiring a globally specified human route
\cite{gao2025large,moustris2026shared}. The problem becomes
interaction-critical, however, when a route segment remains traversable for
the operator but infeasible for the robot--payload system. The robot must then
depart locally from the indicated path while preserving proximity, service
continuity, and task progress. The central challenge is therefore not merely
collision-free following, but maintaining collaboration while task-level route
guidance and executable robot motion temporarily disagree. Figure~\ref{fig:realworld} shows the physical human-guided transport setting considered in this work, in which the robot must reconcile operator-provided route guidance with its own local executability.

The operator's motion, however, cannot be treated as an exogenous reference
trajectory. Interactive navigation has long recognized that human and robot
motions can evolve as coupled rather than independent processes
\cite{trautman2010unfreezing,hirose2024sacson}. In human-guided transport, this coupling has a particular structure. The operator's direction
carries a persistent route tendency, whereas short-horizon speed and
locomotion state may adapt to the robot's realized progress. Recent methods across collaborative
interaction and social navigation make this dependence increasingly explicit by
conditioning predicted human behavior or future scene evolution on contemplated
robot actions
\cite{kedia2024interact,samavi2025sicnav,hu2026navthinker}. Consequently, from
the same observed interaction history, different executable robot commands may
cause the operator to continue, slow down, wait, or resume, producing different
joint human--robot transitions. A controller that ignores this action dependence
may select a maneuver that is feasible for the robot in isolation yet causes
unnecessary operator waiting, excessive separation, or stalled collaborative
progress.

Human-aware navigation has developed rich mechanisms for representing local
geometry, social context, personal space, and interaction structure
\cite{kruse2013human,rodriguezcriado2024_sngnn2dv2}. Person-following and
memory-driven navigation further exploit recent observations to infer guidance
cues and latent motion dynamics under partial observability
\cite{pang2020_hsrl_person_following,montero2025_memory_driven_drl}. These
representations provide increasingly informative estimates of how an observed
interaction may evolve. At a given decision step, however, observation history
is common to all robot candidates. A human continuation inferred primarily from
that shared history does not by itself distinguish responses caused by choosing
different candidate robot actions. Under route--feasibility mismatch, the
maneuver being evaluated may itself alter the operator's pace or waiting
response. Comparing several robot candidates against the same human future can
therefore pair some actions with responses that those actions would not induce.
The relevant object of comparison is consequently not a robot command alone,
but the joint human--robot transition induced by that command.

Action-conditioned prediction still does not ensure a valid comparison when
the conditioned action is not the command that ultimately enters the physical
interaction loop. Control-barrier-function formulations can project nominal controls into
safety-admissible sets \cite{ames2016control}, while safety layers and
safe-learning architectures may project, modify, reject, or replace nominal
actions before physical execution
\cite{dalal2018safe_exploration,brunke2022_safe_learning_review}. Because the
operator responds to the realized robot motion rather than to the
pre-realization proposal, predicting the operator under one action while
rolling out, evaluating, or executing another produces an internally mismatched
joint transition. We refer to the required alignment as
\emph{action--response consistency}. It has two parts. Candidate specificity
requires every executable candidate to be paired with the operator response
expected under that candidate. Realization consistency requires response
prediction, joint rollout, candidate evaluation, and physical execution to
refer to the same executable command.

CARE operationalizes action--response consistency by maintaining a common executable-action index from candidate realization through operator-response prediction, interaction evaluation, and execution. A frozen nominal policy supplies the initial motion proposal, but proposal generation is separated from the final interaction decision. For
the nominal proposal and each bounded local alternative, the same
deployment-time realization mapping first determines the command that can
actually enter the human--robot loop. The resulting executable action then
indexes the route-aligned operator-response prediction, the matched
short-horizon robot--operator rollout, and the fixed joint interaction utility
used for candidate comparison. An alternative is accepted only when its matched
transition improves predicted operator-motion continuity, service coordination,
and local feasibility relative to the realized nominal action. CARE therefore compares and refines the matched interaction consequences of executable actions rather than correcting robot commands against a human forecast held fixed across candidates.

Bounded reciprocal refinement assumes that a coordination-preserving
executable action remains available within a local neighborhood of the
current base proposal. Persistent loss of progress provides evidence that
this local neighborhood has become unproductive. CARE therefore treats
progress as a proposal-regime variable. Under routine interaction, the
frozen PPO policy provides the base proposal for reciprocal refinement.
Under sustained mobility loss, a progress-oriented DWA proposal temporarily
replaces the PPO proposal as the search center, while the same downstream
action realization, operator-response prediction, matched rollout, and
interaction evaluation are retained. Once the low-mobility condition clears, the PPO policy again provides the base proposal.

The main contributions of this work are summarized as follows:
\begin{enumerate}
\item We formulate local decision making under route--feasibility mismatch
over executable-action-indexed joint human--robot transitions. This
formulation establishes action--response consistency, comprising
candidate specificity and realization consistency, as the requirement for
valid reciprocal candidate evaluation when operator behavior depends on the
robot motion that is actually executed.

\item We introduce CARE, a closed-loop controller that maintains
action--response consistency from candidate realization to execution.
Each candidate is made executable before conditioning operator-response
prediction, matched joint rollout, and interaction evaluation. Bounded
reciprocal refinement regulates transient route--feasibility mismatch,
while persistent progress loss temporarily replaces the nominal PPO base
with a progress-oriented DWA proposal and re-centers the same reciprocal
evaluation process.

\end{enumerate}

\section{Related Work}
\label{sec:related}

\subsection{Human-Aware Navigation and Human-Guided Transport}

Human-aware navigation extends geometric collision avoidance toward robot
behavior that accounts for social interaction, comfort, predictability, and
human motion. Early and recent surveys organize this literature around
human-aware planning, interaction modeling, social conventions, and navigation
under uncertainty
\cite{kruse2013human,singamaneni2024_socialnav_survey,
mavrogiannis2023_core_challenges_social_nav}. Human-following and
human-guidance mobile robots introduce a related but distinct interaction
structure in which human motion can directly convey task-level navigation
information \cite{eirale2025human}. Human-in-the-loop navigation frameworks
likewise combine human perception or guidance with robot-side decision making
and action realization, allowing the effective division of navigation authority
to depend on the interaction context \cite{bozorgi2023beyond}.

Within person-following systems, sensing and tracking methods estimate human
motion from onboard perception and provide the geometric and behavioral cues
required for local following. Multisensor fusion, for example, combines ranging
and visual information to improve person tracking under realistic sensing
conditions \cite{sarmento2024fusion}. Socially aware navigation instead focuses
more broadly on how robot motion should respect interaction structure.
Inverse-reinforcement-learning methods have learned socially compliant
navigation costs from demonstrations
\cite{kretzschmar2016socially}, while prediction-aware local planners optimize
trajectories with respect to anticipated human motion
\cite{perez2025social}. Recent reviews of deep-reinforcement-learning navigation
similarly emphasize the role of interaction representations and adaptive local
decision making in crowded environments \cite{le2024comprehensive}.

These lines of work provide important mechanisms for perceiving humans,
representing social structure, and adapting robot motion. Human-guided
transport, however, places the operator in a dual role that is central to the
present problem. The operator is both a source of persistent task-level route
guidance and an interaction partner whose short-horizon pace can change in
response to robot execution. Preserving the first role while allowing the
second to vary across robot candidates motivates a more explicit treatment of
candidate-dependent joint transitions.

\subsection{Action-Conditioned Prediction and Reciprocal Planning}

Reciprocal navigation treats human and robot motions as coupled processes rather
than independently evolving trajectories. Reciprocal velocity-obstacle methods
distribute collision-avoidance responsibility among interacting agents
\cite{vanDenBerg2011reciprocal}, while pedestrian-friendly navigation methods
modify robot motion to avoid freezing and related crowd-interaction effects
\cite{sathyamoorthy2020_frozone}. Learning-based approaches additionally encode
group cohesion and social relations when selecting robot motion
\cite{sathyamoorthy2022_comet}. Game-theoretic formulations further represent
human--robot interaction through coupled strategic dynamics
\cite{galati2025game}. Other approaches optimize robot behavior with respect to
the disturbance it induces in surrounding human motion
\cite{mozzarelli2024_socially_aware_dwa}.

Reciprocity at the motion-interaction level, however, does not automatically
provide a candidate-specific model of how a task-guiding operator will respond
to each contemplated robot command. Interactive prediction methods make this
dependence explicit. InteRACT conditions human intent prediction on planned
robot actions \cite{kedia2024interact}, whereas SICNav jointly reasons about
robot motion and predicted crowd behavior within an interactive optimization
framework \cite{samavi2025sicnav}. Prediction-aware social planning has also
been extended with diffusion-based human trajectory models that couple future
crowd evolution with safe robot decision making
\cite{samavi2025sicnavdiffusion}.

These methods establish that different contemplated robot actions may
correspond to different human futures. The deployment setting
considered here adds another stage to this relationship. A contemplated action
may be modified by feasibility or safety processing before entering the
physical interaction loop. Candidate-specific prediction is therefore
insufficient if prediction is indexed by the pre-realization proposal while
rollout or execution uses a different command. The gap addressed by CARE is
thus not whether robot actions affect human behavior, but how this dependence
remains aligned with the executable command throughout prediction, evaluation,
and execution.

\subsection{Action Realization and Deployment-Time Refinement}

Local navigation methods produce robot-side motion commands through a wide
range of planning and learning paradigms. Dynamic-window approaches search
locally admissible velocity spaces and have been extended toward longer-horizon
and socially aware navigation
\cite{jian2023_ltdwa,mozzarelli2024_socially_aware_dwa}. Human-aware local
planners incorporate social objectives into local motion generation
\cite{kivrak2021_cpsfm_socialnav}. Constrained-optimization methods formulate
short-horizon navigation under uncertainty and interaction objectives
\cite{akhtyamov2025_constrained_optimization}. Learned navigation policies
provide another source of nominal motion proposals, including actor--critic
controllers and velocity-obstacle-guided reinforcement learning
\cite{jiang2022itd3cln,xie2023_drl_vo_tro}. Despite their different internal
representations, these approaches share a deployment interface in which a
robot-side command or trajectory proposal is produced from the current state.

The proposal produced by a planner or learned policy is not necessarily the
command ultimately sent to the platform. Control-barrier-function controllers
can project nominal inputs into safety-admissible control sets
\cite{ames2016control}, while safe-learning architectures employ filtering,
shielding, or corrective mechanisms between policy output and physical
execution \cite{brunke2022_safe_learning_review}. Such transformations are
normally introduced to enforce robot-side feasibility or safety. In an
interactive transport task, however, modifying the robot command also changes
the realized motion observed by the operator and can therefore alter the
resulting human--robot transition.

Deployment-time refinement provides a complementary mechanism for improving a
fixed or pretrained controller without replacing its underlying policy.
Residual-policy methods learn additive corrections on top of an existing
controller
\cite{johannink2019_residual_rl,silver2018_residual_policy}, whereas more recent
inference-time approaches use predictive models to evaluate or improve action
proposals according to their anticipated future consequences
\cite{qi2026inference}. These approaches demonstrate that substantial behavior
improvement can be obtained between nominal policy inference and physical
execution.

For human-guided transport, the relevant deployment consequence is not solely
the corrected robot action. When the operator adapts to robot execution, a
candidate modification simultaneously changes the human response against which
that candidate should be evaluated. CARE therefore connects interactive
prediction, action realization, and deployment-time refinement through the executable robot action as a common decision index. For each candidate, the same realized command determines the predicted operator response, robot rollout, collaborative evaluation, and action sent to the platform. Candidate comparison is consequently performed over matched human--robot transitions rather than robot commands paired with a shared or pre-realization human forecast.

\section{Method}
\label{sec:method}

During reciprocal operation, CARE evaluates each robot proposal through
the command that can actually enter the human--robot interaction. Let
\(h_t\) denote the available interaction history, \(\mathbf o_t\) the
current local interaction observation, including operator-relative state,
observed operator velocity, and visible obstacle geometry, and
\(\mathbf v\in\mathcal V_t\) a local robot-motion proposal. The reciprocal decision principle is
\begin{equation}
\begin{aligned}
\mathbf u_t(\mathbf v)
&=
\mathcal R_t(\mathbf v;\mathbf o_t),
\\
\mathcal U_t^{\mathrm{exec}}(\mathbf o_t)
&=
\left\{
\mathbf u_t(\mathbf v)
:
\mathbf v\in\mathcal V_t
\right\},
\\
\mathbf u_t^\star
&\in
\arg\max_{\mathbf u\in
\mathcal U_t^{\mathrm{exec}}(\mathbf o_t)}
J_t(\mathbf u).
\end{aligned}
\label{eq:care_core}
\end{equation}
Here, \(J_t(\mathbf u)\) is the matched executable-action score
instantiated in Sec.~\ref{sec:interaction_utility}: operator-response
prediction, relative human--robot rollout, and candidate evaluation are
all computed under the same executable command \(\mathbf u\).
Equation~\eqref{eq:care_core} states the decision principle over the
executable image of a proposal set, whereas
Sec.~\ref{sec:bounded_refinement} instantiates this set as a finite
bounded candidate set and applies the improvement-margin acceptance rule.

\begin{figure*}[t]
    \centering
    \includegraphics[width=\textwidth]{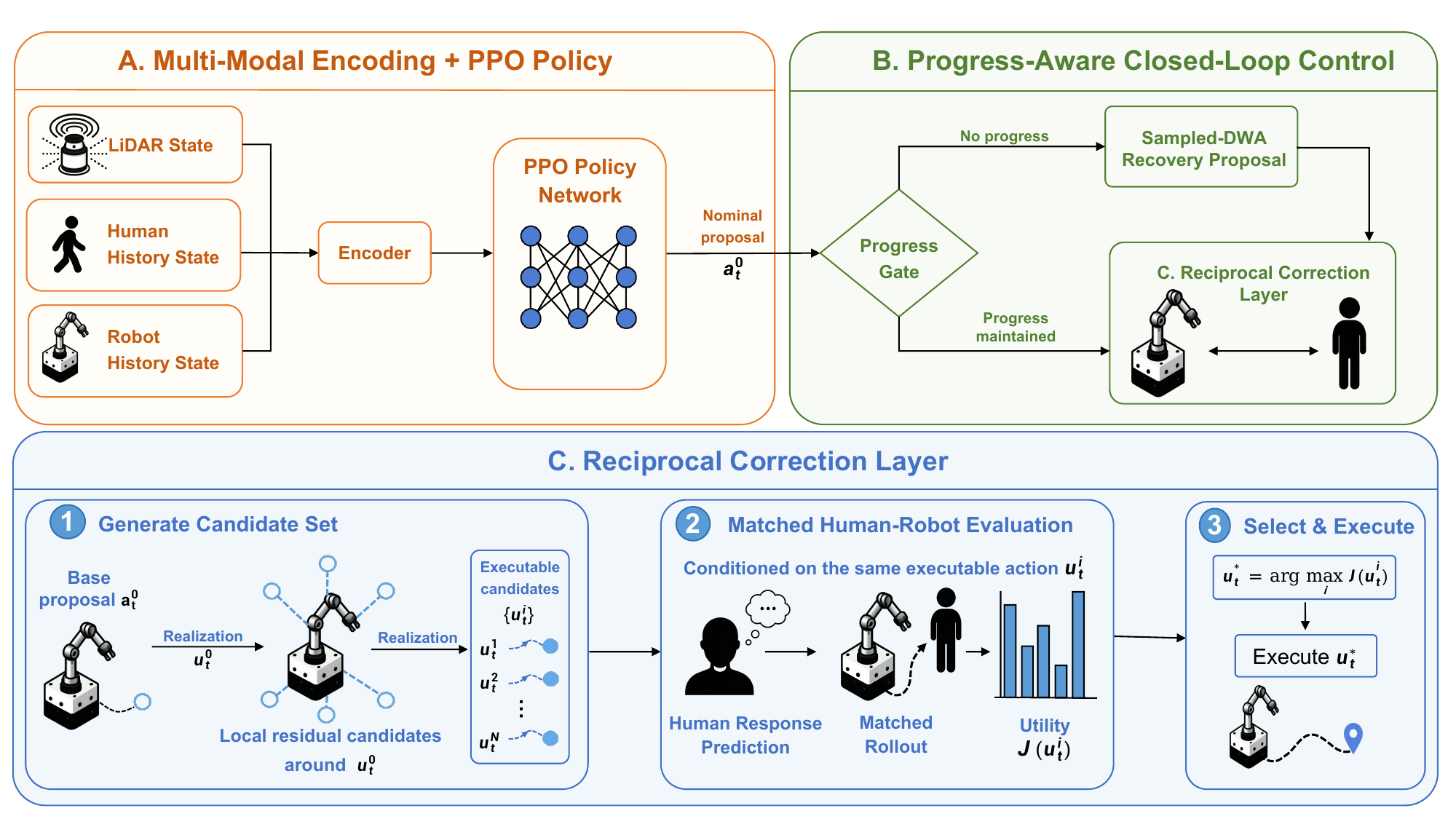}
    \caption{Schematic overview of the CARE closed loop. The progress
    supervisor selects the nominal or recovery proposal, after which
    candidate actions undergo realization, operator-response prediction,
    matched rollout, and interaction evaluation before execution.
    Simplified action notation is used for readability; the formal action
    realization is defined in Sec.~\ref{sec:action_realization}, while the
    candidate construction and improvement-margin selection rules are given
    in Eqs.~\eqref{eq:realized_base}--\eqref{eq:final_action}.}
    \label{fig:care_framework}
\end{figure*}

Figure~\ref{fig:care_framework} provides a schematic overview of the
CARE architecture, from nominal proposal generation to progress-aware
recovery and executable-action-conditioned reciprocal evaluation.
The figure uses simplified action notation for readability; the formal
proposal, realization, and execution variables are defined below.

The architecture therefore separates three roles: the nominal or recovery
policy determines where local search starts, action realization determines
what motion can physically enter the interaction, and the reciprocal layer
compares the matched human--robot consequences of those executable actions.

A frozen nominal policy normally supplies the base motion proposal but does
not directly determine the executed interaction. A progress-aware supervisor,
defined in Section~\ref{sec:progress_recovery}, monitors whether the local
proposal neighborhood continues to support mobility. Let \(g_t^{\mathrm{rec}}\) denote
its recovery indicator. The base proposal is

\begin{equation}
\mathbf v_t^{\mathrm b} =
\begin{cases}
\mathbf v_t^{\mathrm{DWA}}, & g_t^{\mathrm{rec}} = 1,\\
\mathbf v_t^{\pi}, & g_t^{\mathrm{rec}} = 0,
\end{cases}
\label{eq:base_proposal}
\end{equation}

where \(\mathbf v_t^{\mathrm{DWA}}\) is the recovery proposal defined in
Section~\ref{sec:progress_recovery}. Thus, recovery changes the source and search
center of the proposal rather than bypassing the reciprocal evaluator.

The resulting base proposal and all bounded local alternatives are processed
through the same action-realization map before operator-response prediction,
matched rollout, and interaction scoring. Persistent low mobility therefore
changes where CARE searches from, whereas the executable-action-conditioned
evaluation principle remains unchanged.

\subsection{Executable-Action-Indexed Decision Formulation}
\label{sec:decision_formulation}

Planar human-guided collaborative transport is considered with a holonomic
omnidirectional mobile robot. The operator communicates the task-level
route through motion and may transition among Walk, Slow, Wait, and Resume
in response to the robot motion that is actually executed. The robot
receives neither the global route nor a complete global obstacle map;
local guidance must therefore be inferred from the observed interaction.

Let the operator position relative to the robot be
\(\mathbf r_t=\mathbf p_t^H-\mathbf p_t^R\), let
\(\mathbf v_t^H\) be the observed operator velocity, and let
\(d_t=\|\mathbf r_t\|_2\). The nearest visible obstacle-surface clearance is
\(c_t\), with \(\mathbf r_t^{\mathrm{obs}}\) denoting the corresponding vector
from the robot center. For an executed translational command
\(\mathbf u_t=[v_{x,t},v_{y,t}]^\top\), the one-step robot motion is
\(\mathbf p^R_{t+1}=\mathbf p^R_t+\Delta t\,\mathbf u_t\), with the
component-wise bound
\(\mathbf u_t\in\mathcal A=[-v_{\max},v_{\max}]^2\).

The reciprocal layer controls planar translational velocity.
Here, \emph{executable} refers to a planar command satisfying the modeled
velocity and local interaction constraints.

A shared-response formulation would evaluate all robot candidates against one
operator continuation \(\widehat{\mathbf Y}^{H}_t(h_t)\) inferred from the
common observation history. This is appropriate only when the near-term
operator response is effectively independent of the candidate robot action.
Conditioning instead on the raw proposal,
\(\widehat{\mathbf Y}^{H}_t(h_t,\mathbf v)\), represents action dependence but
remains internally mismatched whenever
\(\mathcal R_t(\mathbf v;\mathbf o_t)\neq\mathbf v\), because the operator is
exposed to the realized command rather than the proposal before constraint
processing.

A valid reciprocal comparison therefore requires two forms of consistency.
\emph{Candidate specificity} requires each distinct executable candidate to be
paired with the operator response predicted under that candidate.
\emph{Realization consistency} requires operator-response prediction, relative
human--robot rollout, interaction evaluation, and first-step execution to refer
to the same post-realization command.

The nominal proposal is produced by a frozen PPO actor
\cite{schulman2017proximal}. Its 100-D observation comprises 36 ordered
LiDAR sectors, a 32-D encoding of the current 9-D robot--operator state,
and a 32-D GRU encoding of a 16-step interaction history. The current
state contains operator velocity, relative position, the previous executed
action, and local obstacle geometry. Each history step contains relative
position, operator velocity, the PPO proposal, the executed action, and
an intervention indicator. The actor produces
\[
\mathbf v_t^\pi
=
v_{\max}\pi_\theta(\mathbf z_t^\pi),
\qquad
\pi_\theta(\cdot)\in[-1,1]^2.
\]
Under nominal operation, \(\mathbf v_t^\pi\) provides the base proposal for
reciprocal evaluation. During sustained progress loss, the recovery supervisor
temporarily replaces this base with \(\mathbf v_t^{\mathrm{DWA}}\), while
retaining the same downstream action realization, operator-response
prediction, matched rollout, and interaction evaluation.

\subsubsection{Executable Action Realization}
\label{sec:action_realization}

CARE realizes every robot proposal before interaction evaluation.
Physical velocity limits and state-dependent operator or obstacle
constraints may modify the proposed command, so downstream prediction
and scoring are indexed by the resulting executable action.

The realization map first clips the proposal to the component-wise velocity
bounds and then processes active operator-distance and visible-obstacle
constraints. For the operator constraint, let
\(\mathbf n_t^H=\mathbf r_t/d_t\). For a valid LiDAR return \(r\), let
\(d_{t,r}\) be its range, \(r_R\) the effective planar footprint radius used
by the controller,
\(c_{t,r}=d_{t,r}-r_R\) the corresponding surface clearance, and
\(\mathbf n_{t,r}\) its unit direction. Define the operator-distance and
obstacle-clearance barrier functions as
\(h_H=d_t-d_H\) and
\(h_{O,r}=c_{t,r}-c_{\mathrm{hard}}\), respectively.
Applying the local barrier condition
\(\dot h+k h\geq0\)~\cite{ames2016control} to each active constraint yields
\begin{equation}
\begin{aligned}
    (\mathbf n_t^H)^\top\mathbf u
    &\leq
    (\mathbf n_t^H)^\top\mathbf v_t^H
    -k_H(d_H-d_t),
    && 0<d_t<d_H,
    \\
    \mathbf n_{t,r}^{\top}\mathbf u
    &\leq
    k_O(c_{t,r}-c_{\mathrm{hard}}),
    && 0\leq c_{t,r}<d_O.
\end{aligned}
\label{eq:realization_constraints}
\end{equation}

The first row limits robot closing motion relative to the moving operator,
whereas the second reduces obstacle-closing velocity as a visible surface
approaches the hard-clearance margin. Unknown returns are excluded, and
neighboring active rays are grouped into the representative surface normals
used by the controller.

Starting from the clipped proposal, the active half spaces are processed
cyclically for \(K_{\mathrm{cyc}}\) passes. A satisfied row leaves the current
command unchanged, whereas a violated row moves it orthogonally to the
corresponding half-space boundary before the component-wise action bounds are
re-applied.

If the active rows remain mutually inconsistent after the configured passes,
obstacle constraints are retained before the operator-distance constraint,
followed when necessary by the configured obstacle-only non-closing
relaxation. The complete deterministic procedure defines
\[
    \mathbf u_t(\mathbf v)
    =
    \mathcal R_t(\mathbf v;\mathbf o_t)
\]
and is applied identically to nominal, reciprocal, and recovery proposals.
The bounded cyclic half-space procedure defines the action-realization map
\(\mathcal R_t\).

\subsubsection{Route-Persistent, Action-Responsive Operator Model}
\label{sec:operator_response_model}

CARE separates the operator information that remains common across candidate robot actions from the behavior allowed to vary with the executable command. The filtered operator heading \(\mathbf q_t\) represents the locally persistent
route tendency and is retained while the operator is nearly stationary. The heading \(\mathbf q_t\) is estimated solely from observed operator motion and does not use the hidden route used to construct the episode. Operator speed and locomotion mode may instead vary with the route-aligned effect of the realized robot action. The heading filter is fixed before evaluation. For each executable candidate, the response model receives the eight-step history
\[
    \mathcal H_t^H
    =
    \left\{
        [\mathbf r_i^\top,(\mathbf v_i^H)^\top]^\top
    \right\}_{i=t-7}^{t}
\]
and the current geometry
\[
    \boldsymbol\gamma_t
    =
    [c_t,(\mathbf r_t^{\mathrm{obs}})^\top,
      \ell_t,e_t^\perp]^\top,
\]
where
\(\ell_t=\mathbf q_t^\top\mathbf r_t-d_f\) and
\(e_t^\perp=(\mathbf q_t^\perp)^\top\mathbf r_t\).
Here, \(\mathbf q_t^\perp\) denotes the \(90^\circ\) rotation of
\(\mathbf q_t\). These variables are standardized using statistics from the
response-model training split.

The action-dependent conditioning variable is the route-aligned lag
increment
\(\chi_t^\parallel(\mathbf u)
=-\Delta t\,\mathbf q_t^\top\mathbf u\).
The deployed response model embeds this scalar as
\(\mathbf e_t(\mathbf u)
=[\chi_t^\parallel(\mathbf u),\,0]^\top\), so only the first channel varies
across candidate actions. The frozen model predicts
\begin{equation}
    \widehat{\mathbf Y}^{H}_t(\mathbf u)
    =
    F_\phi
    \left(
        \mathcal H_t^H,
        \boldsymbol\gamma_t,
        \mathbf e_t(\mathbf u)
    \right).
    \label{eq:response_prediction}
\end{equation}
The model produces five predictions at \(0.2\,\mathrm{s}\) intervals over a
\(1\,\mathrm{s}\) horizon, including operator speed and the probabilities of
Walk, Slow, Wait, and Resume.

The effective action dependence is therefore one-dimensional and
route-aligned. Candidates with the same
\(\mathbf q_t^\top\mathbf u\) receive the same predicted operator response,
while their lateral consequences remain distinguishable through the matched
relative rollout.

Let \(\tau_k=k\Delta\tau\). The predicted operator displacement and matched
relative future are
\begin{equation}
\begin{aligned}
    \widehat{\Delta\mathbf p}^{H}_{t,k}(\mathbf u)
    &=
    \mathbf q_t
    \sum_{j=1}^{k}
        \Delta\tau\,
        \widehat s^{H}_{t,j}(\mathbf u),
    \\
    \widehat{\mathbf r}_{t,k}(\mathbf u)
    &=
    \mathbf r_t
    +
    \widehat{\Delta\mathbf p}^{H}_{t,k}(\mathbf u)
    -
    \tau_k\mathbf u.
\end{aligned}
\label{eq:matched_rollout}
\end{equation}
Thus, response prediction, robot propagation, and relative-state evaluation
all refer to the same executable command. The constant-command rollout is used
only as a local evaluation model; the controller executes its first command
and replans after the next observation.

The response model is trained offline on episode-disjoint interaction
rollouts using post-realization executed commands as conditioning actions.
Speed predictions are supervised with a Smooth-\(L_1\) loss, and an auxiliary
Smooth-\(L_1\) term is applied to the displacement obtained by integrating the
predicted speed sequence. Walk, Slow, Wait, and Resume probabilities are
supervised with weighted cross-entropy. The checkpoint is selected using
validation displacement error and is frozen before controller evaluation.

\subsubsection{Matched Interaction Utility}
\label{sec:interaction_utility}

Each executable candidate is evaluated from operator-motion continuity,
predicted waiting, service coordination, and local interaction risk.
A command-continuity term penalizes abrupt changes from the previously
executed action. Define the terminal route-aligned lag as
\(\widehat\ell_{t,N}(\mathbf u)
=\mathbf q_t^\top\widehat{\mathbf r}_{t,N}(\mathbf u)-d_f\).
The matched-transition terms are
\begin{equation}
\begin{aligned}
    M_t^H(\mathbf u)
    &=
    \left\|
        \widehat{\Delta\mathbf p}^{H}_{t,N}(\mathbf u)
    \right\|_2,
    \\
    W_t(\mathbf u)
    &=
    \frac{1}{N}
    \sum_{k=1}^{N}
    \widehat p^H_{t,k}(\mathrm{Wait}\mid\mathbf u),
    \\
    C_t(\mathbf u)
    &=
    \left|\widehat\ell_{t,N}(\mathbf u)\right|
    +
    \lambda_s
    \left|
        \left\|\widehat{\mathbf r}_{t,N}(\mathbf u)\right\|_2-d_f
    \right|
    \\
    &\quad+
    \lambda_{\mathrm{tr}}
    \left[
        \left\|\widehat{\mathbf r}_{t,N}(\mathbf u)\right\|_2-d_{\max}
    \right]_+,
    \\
    \Lambda_t(\mathbf u)
    &=
    \lambda_H
    \left[
        d_{\min}^{H}
        -
        \left\|\widehat{\mathbf r}_{t,N}(\mathbf u)\right\|_2
    \right]_+
    \\
    &\quad+
    \lambda_O
    [\mathbf u^\top\mathbf n_t^{\mathrm{obs}}]_+
    [c_{\mathrm{risk}}-c_t]_+.
\end{aligned}
\label{eq:utility_components}
\end{equation}

Here, \(d_f\) is the desired following distance and \(d_{\max}\) is the
service-range limit. The coordination term penalizes terminal route-aligned
lag, radial following-distance error, and separation beyond the service range.
The soft margin \(d_{\min}^{H}\) penalizes excessive human--robot proximity,
whereas \(c_{\mathrm{risk}}\) activates the obstacle-closing penalty. For a
valid nearest-obstacle return,
\(\mathbf n_t^{\mathrm{obs}}
=\mathbf r_t^{\mathrm{obs}}/
\|\mathbf r_t^{\mathrm{obs}}\|_2\); the obstacle component is zero when no valid
return exists.

The reciprocal interaction score is
\begin{equation}
\begin{aligned}
    J_t(\mathbf u)
    ={}&
    w_M M_t^H(\mathbf u)
    -
    w_W W_t(\mathbf u)
    -
    C_t(\mathbf u)
    \\
    &-
    \Lambda_t(\mathbf u)
    -
    w_\Delta
    \|\mathbf u-\mathbf u_{t-1}\|_2.
\end{aligned}
\label{eq:interaction_score}
\end{equation}
The terms respectively reward short-horizon operator-motion continuity and
penalize predicted waiting, service-configuration error, local interaction
risk, and abrupt command changes. All terms are computed from the same
executable action and its matched operator prediction. The fixed score evaluates the local matched interaction induced by each
executable action.

\subsection{Bounded Reciprocal Refinement}
\label{sec:bounded_refinement}

The selected base proposal is first realized as

\begin{equation}
\mathbf u_t^0 =
\mathcal R_t(\mathbf v_t^{\mathrm b};\mathbf o_t),
\label{eq:realized_base}
\end{equation}

with matched score
\(
J_t^0 = J_t(\mathbf u_t^0).
\)
Its operator response, relative rollout, and interaction score are always
evaluated under this realized base command.

Bounded candidate search is activated when recovery is active or when at
least one of three reciprocal conditions holds:
\(c_t<\tau_c\),
\(\left|\widehat{\ell}_{t,N}(\mathbf u_t^0)\right|>\tau_\ell\), or
\(W_t(\mathbf u_t^0)>\tau_W\). Thus, a recovery proposal always serves as the
center of a bounded reciprocal search, whereas under nominal operation the
same search is activated only when the reciprocal gate opens. Every generated
candidate is ranked by the same executable-action score in
Eq.~\eqref{eq:interaction_score}.

\subsubsection{Local Search over the Deployed Interaction Map}
The realized base command is reused as the center of the local proposal
search. Let \(\mathbf e_i\) be the \(i\)-th coordinate vector,
\(i\in\{1,2\}\). CARE constructs a score-informed direction from symmetric
probes of the complete post-realization evaluation map:

\begin{equation}
[\mathbf g_t^{\mathrm{fd}}]_i
=
\frac{1}{2\epsilon}
\left[
J_t\!\left(
\mathcal R_t(\mathbf u_t^0+\epsilon\mathbf e_i;\mathbf o_t)
\right)
-
J_t\!\left(
\mathcal R_t(\mathbf u_t^0-\epsilon\mathbf e_i;\mathbf o_t)
\right)
\right].
\label{eq:finite_difference_direction}
\end{equation}

Each probe is independently re-realized, assigned a fresh
executable-action-conditioned operator prediction, rolled out, and scored.
Thus, \(\mathbf g_t^{\mathrm{fd}}\) reflects local variation of the complete
deployment map rather than only the response network. The finite-difference search operates directly on the deployed post-realization evaluation map.

In addition to the score-informed direction, CARE constructs a compact
geometry-aware search basis. When
\(\|\mathbf u_t^0\|_2>\tau_u\), the active-motion direction is the normalized
realized base command
\(\mathbf u_t^0/\|\mathbf u_t^0\|_2\). The service-restoration direction
\(\mathbf d_t^{\mathrm{srv}}\) is
\(\mathbf r_t/d_t\) when \(d_t>d_f\) and
\(-\mathbf r_t/d_t\) when \(0<d_t<d_f\); it is omitted when the distance error
falls below the configured numerical tolerance.

For a valid nearest-obstacle normal
\(\mathbf n_t^{\mathrm{obs}}
=[n_{t,x}^{\mathrm{obs}},n_{t,y}^{\mathrm{obs}}]^\top\),
the two obstacle tangents are
\(\mathbf t_t^\pm
=\pm[-n_{t,y}^{\mathrm{obs}},n_{t,x}^{\mathrm{obs}}]^\top\).
The same filtered route heading \(\mathbf q_t\) used by the operator-response
model is also used to construct the guide--tangent directions by normalizing
\(\mathbf q_t+\lambda_b\mathbf t_t^\pm\).

For each tangent \(\mathbf t\), the visible-free-space score
\(F_t^{\mathrm{free}}(\mathbf t)\) is the mean of valid LiDAR ranges
clipped to \([0,3]\,\mathrm{m}\) within a \(25^\circ\) angular half-window
centered on \(\mathbf t\). CARE retains
\[
    \mathbf d_t^{\mathrm{tan}}
    \in
    \arg\max_{\mathbf t\in\{\mathbf t_t^+,\mathbf t_t^-\}}
    F_t^{\mathrm{free}}(\mathbf t).
\]
The obstacle influence is
\(\beta_t=
\operatorname{clip}((c_{\mathrm{blend}}-c_t)/
(c_{\mathrm{blend}}-c_{\mathrm{hard}}),0,1)\)
for a valid obstacle return and zero otherwise. The corresponding
obstacle-aware service direction is obtained by normalizing
\((1-\beta_t)\mathbf d_t^{\mathrm{srv}}
+\lambda_{\mathrm{tan}}\beta_t\mathbf d_t^{\mathrm{tan}}\).

The clearance-scaled residual radius is
\(\rho_t=
\rho\,\operatorname{clip}((c_t-c_{\mathrm{hard}})/
(c_{\mathrm{search}}-c_{\mathrm{hard}}),0,1)\)
for a valid obstacle return and \(\rho_t=\rho\) otherwise. The final direction
set \(\mathcal D_t\) contains valid score-informed, active-motion,
service-restoration, obstacle-aware service, obstacle-tangent, and
guide--tangent directions. The score-informed direction is omitted when
\(\|\mathbf g_t^{\mathrm{fd}}\|_2<\tau_{\mathrm{fd}}\); directions with
negligible norm are removed, duplicates are merged using the configured
angular tolerance, and obstacle-dependent directions are excluded when no
valid obstacle normal is available.

Let \(\mathcal S\) denote the fixed set of residual scales and
\(\mathcal D_t\) the valid search directions. When reciprocal search is active,
the executable candidate set is
\begin{equation}
\begin{aligned}
    \mathcal U_t^{\mathrm{cand}}
    =
    \{\mathbf u_t^0\}
    \cup
    \big\{
        \mathcal R_t(
            \mathbf u_t^0+\rho_t s\mathbf d;
            \mathbf o_t
        )
        :\,
        \mathbf d\in\mathcal D_t,\;
        s\in\mathcal S
    \big\}.
\end{aligned}
\label{eq:candidate_set}
\end{equation}
When neither recovery nor the reciprocal gate activates bounded search,
the set reduces to \(\{\mathbf u_t^0\}\). Duplicate post-realization commands are removed before
response prediction and scoring.

The residual is centered on the realized base action rather than directly on
its pre-realization proposal. Under nominal operation this base originates
from the frozen PPO actor; under recovery it originates from the selected
DWA proposal. Every local alternative is independently re-realized because
the perturbation may activate constraints that were inactive for the base
command.

For the realized base command \(\mathbf u_t^0\) and its bounded executable
candidate set \(\mathcal U_t^{\mathrm{cand}}\), define

\begin{equation}
\mathbf u_t^\star
\in
\arg\max_{\mathbf u\in\mathcal U_t^{\mathrm{cand}}}
J_t(\mathbf u).
\end{equation}

The final executed command is

\begin{equation}
\mathbf u_t =
\begin{cases}
\mathbf u_t^\star,
&
J_t(\mathbf u_t^\star)
>
J_t(\mathbf u_t^0)+\eta,
\\[2mm]
\mathbf u_t^0,
&
\text{otherwise}.
\end{cases}
\label{eq:final_action}
\end{equation}

Thus, the recovery supervisor determines the source of the base proposal,
whereas final action selection remains governed by the same matched
executable-action score. Under nominal operation, reciprocal refinement is
centered on the realized PPO proposal; under recovery, it is re-centered on
the realized DWA proposal.

\subsection{Progress-Aware Recovery}
\label{sec:progress_recovery}

Bounded refinement assumes that a useful executable alternative remains
near the current base proposal. CARE monitors persistent low mobility
using operator lag, nominal approach motion, and obstacle-proximal
low-speed motion. Specifically, a step is marked as low mobility when
any of these fixed-threshold progress conditions is satisfied; all
thresholds are fixed before evaluation. Recovery is triggered after
\(L_{\mathrm{rec}}=8\) consecutive low-mobility steps and remains active
until the low-mobility condition clears.

During recovery, a sampled DWA planner~\cite{fox1997dynamic} provides a progress-oriented base
proposal \(\mathbf v_t^{\mathrm{DWA}}\). This proposal replaces the
nominal PPO base but is otherwise treated identically: it undergoes the
same action realization, bounded candidate generation,
executable-action-conditioned response prediction, matched rollout, and
interaction evaluation. Once mobility recovers, the PPO policy again
provides the base proposal.

\section{Experiments}
\label{sec:experiments}

\subsection{Evaluation Scenarios and Protocol}
\label{sec:evaluation_protocol}

We evaluate single-operator planar collaborative transport in a custom
2-D environment matched to the geometry and \(20\,\mathrm{Hz}\) control
interface of the physical robot--payload platform. Learned components are trained on a training-map pool under the Side and
Route conditions and are frozen before evaluation. Generalization is
evaluated on ten corridor maps and independently generated routes excluded
from training and checkpoint selection.

The three conditions control the relationship between operator guidance
and robot executability. Clear contains no injected hazard and serves only
as an auxiliary nominal reference. Side places a human-traversable but
robot-non-traversable hazard adjacent to the operator route, so the observed
route direction remains locally usable by the robot. Route places the same
hazard class directly on the operator route: the operator can continue,
whereas the robot must depart locally from the indicated direction. The
Side--Route contrast therefore isolates whether operator guidance remains
an executable local reference.

The simulated operator is independent of the CARE response predictor and
evolves stochastically among Walk, Slow, Wait, and Resume modes in response
to the robot's realized motion. All controllers are evaluated on
seed-matched episodes with identical maps, routes, initial states, hazard
configurations, operator styles, and random streams. The route used to
construct each episode is never exposed to the controller.

The implementation, trained checkpoints, and frozen evaluation configurations will be made publicly available upon publication.

\subsubsection{Compared Methods}

CARE is compared with MPC~\cite{zeng2021safetycritical},
DWA~\cite{fox1997dynamic}, TEB~\cite{rosmann2012trajectory},
and ORCA~\cite{vanDenBerg2011reciprocal}. All comparison methods,
including the PPO reference, are evaluated with the same
deployment-time CBF-based action-realization layer
\cite{ames2016control}, which enforces common velocity,
operator-distance, and obstacle constraints before execution.

For MPC, DWA, TEB, ORCA, and the PPO reference, the common
CBF-based realization map is applied to the single command proposed at
each control step. CARE applies the same realization map to each bounded
candidate before operator-response prediction, matched rollout, and
interaction evaluation. The PPO reference disables reciprocal refinement
and recovery while retaining the same realization layer. Sampled-DWA is
used within CARE only during recovery to provide a temporary base
proposal.

\subsubsection{Evaluation Metrics and Statistical Reporting}
\label{sec:metrics}

Mission success and service availability characterize task continuity.
A mission succeeds when the operator reaches the route endpoint without
a terminal tracking- or service-loss failure. Temporary service
unavailability is permitted and is reflected in service availability,
defined as the fraction of control steps for which the operator is visible
and remains within the service range. Operator wait is the cumulative time
spent in the Wait state. Comfort violation is the fraction of control steps for which the human--robot distance falls below the personal-space threshold. Executed-command smoothness is characterized by a jerk proxy, defined as the episode-mean norm of the second temporal difference of normalized executed actions divided by \(\Delta t^2\), and reported in normalized action units per second squared (a.u./s$^2$). Controller latency measures the computation time required to produce the final executable command, excluding environment stepping and sensor simulation.

Unless otherwise stated, all simulation studies use 300 seed-matched
episodes per method--condition cell. Binary outcomes are reported as
percentages, while continuous metrics are summarized by episode-level
mean and sample standard deviation. Episode-level percentile-bootstrap
95\% confidence intervals are used for all representation-study
summaries and for continuous outcomes in the sensing-range study.
Matched-seed bootstrap 95\% confidence intervals are used for
response-conditioning mechanism summaries. Two-sided Wilson 95\%
confidence intervals are used for binary rates in the delayed-response
and sensing-range studies. Latency measurements are collected
sequentially on the same Intel Xeon Platinum 8352V CPU configuration.

\subsection{Interaction-Selective Performance across Route--Feasibility}
\label{sec:main_comparison}

Table~\ref{tab:main_results} tests whether CARE's relative advantage is
selective to route--feasibility mismatch rather than to the mere presence
of a robot-non-traversable hazard. In the auxiliary Clear evaluation
(not tabulated), all methods complete every episode on both map splits.
In Side, task completion remains near saturation, particularly on
held-out maps, compressing the differences among controllers when the
operator-indicated direction remains locally executable, even when
nearby geometry restricts the robot's available motion.

\begin{table}[ht]
\centering
\caption{Closed-loop performance on training and held-out Side and Route
maps over 300 seed-matched episodes per method--condition cell.
Continuous metrics report mean \(\pm\) standard deviation.
Best results within each map split and condition are shown in bold.}
\label{tab:main_results}

\fontsize{6.2}{6.8}\selectfont
\setlength{\tabcolsep}{1.0pt}
\renewcommand{\arraystretch}{1.08}

\begin{tabular}{@{}l Y Z Z Z Y Z Z Z@{}}
\toprule
& \multicolumn{4}{c}{\textbf{Side}}
& \multicolumn{4}{c}{\textbf{Route}} \\
\cmidrule(lr){2-5}
\cmidrule(lr){6-9}

\textbf{Method}
& \multicolumn{1}{c}{\textbf{Succ.} $\uparrow$}
& \multicolumn{1}{c}{\textbf{Serv.} $\uparrow$}
& \multicolumn{1}{c}{\textbf{Jerk prox.} $\downarrow$}
& \multicolumn{1}{c}{\textbf{Avg. Lat.} $\downarrow$}
& \multicolumn{1}{c}{\textbf{Succ.} $\uparrow$}
& \multicolumn{1}{c}{\textbf{Serv.} $\uparrow$}
& \multicolumn{1}{c}{\textbf{Jerk prox.} $\downarrow$}
& \multicolumn{1}{c}{\textbf{Avg. Lat.} $\downarrow$} \\

&
\multicolumn{1}{c}{\textbf{(\%)}}
& \multicolumn{1}{c}{\textbf{(\%)}}
& \multicolumn{1}{c}{\textbf{(a.u./s$^2$)}}
& \multicolumn{1}{c}{\textbf{(ms)}}
& \multicolumn{1}{c}{\textbf{(\%)}}
& \multicolumn{1}{c}{\textbf{(\%)}}
& \multicolumn{1}{c}{\textbf{(a.u./s$^2$)}}
& \multicolumn{1}{c}{\textbf{(ms)}} \\
\midrule
\multicolumn{9}{c}{\textbf{Training maps}} \\
\midrule

MPC
& 95.33
& 96.76 \pm 14.74
& \bfseries 2.92 \pm 6.03
& 6.62 \pm 2.93
& 95.67
& 95.89 \pm 14.65
& \bfseries 3.16 \pm 3.36
& 6.63 \pm 2.92 \\

ORCA
& 96.67
& 97.68 \pm 12.56
& 8.00 \pm 1.98
& \bfseries 5.53 \pm 1.75
& 75.00
& 75.89 \pm 32.58
& 6.86 \pm 3.19
& \bfseries 5.69 \pm 1.76 \\

TEB
& 95.67
& 97.01 \pm 13.95
& 14.98 \pm 4.99
& 13.99 \pm 5.90
& 93.00
& 92.47 \pm 17.55
& 17.81 \pm 5.22
& 14.69 \pm 6.22 \\

DWA
& 97.67
& 98.64 \pm 7.14
& 9.47 \pm 1.66
& 18.32 \pm 10.17
& 96.33
& 97.30 \pm 9.12
& 9.71 \pm 1.58
& 20.59 \pm 9.47 \\

\rowcolor{black!8}
\textbf{CARE}
& \bfseries 99.33
& \bfseries 99.01 \pm 6.33
& 5.27 \pm 1.93
& 6.74 \pm 9.66
& \bfseries 98.00
& \bfseries 98.07 \pm 9.17
& 5.74 \pm 2.08
& 6.58 \pm 9.53 \\

\midrule

\multicolumn{9}{c}{\textbf{Held-out maps}} \\
\midrule

MPC
& \bfseries 100.00
& 99.97 \pm 0.50
& \bfseries 1.73 \pm 0.66
& 6.51 \pm 1.91
& 96.67
& 98.33 \pm 4.67
& \bfseries 2.19 \pm 2.04
& 6.52 \pm 1.90 \\

ORCA
& \bfseries 100.00
& \bfseries 100.00 \pm 0.00
& 7.96 \pm 1.56
& \bfseries 5.32 \pm 1.58
& 80.67
& 85.70 \pm 24.43
& 7.37 \pm 2.94
& \bfseries 5.49 \pm 1.64 \\

TEB
& \bfseries 100.00
& \bfseries 100.00 \pm 0.00
& 20.93 \pm 3.93
& 13.84 \pm 5.52
& 92.33
& 92.91 \pm 17.52
& 21.45 \pm 4.82
& 13.93 \pm 5.30 \\

DWA
& \bfseries 100.00
& 99.99 \pm 0.16
& 9.04 \pm 1.53
& 16.03 \pm 7.55
& 97.33
& 98.64 \pm 4.18
& 9.08 \pm 1.49
& 15.71 \pm 6.90 \\

\rowcolor{black!8}
\textbf{CARE}
& \bfseries 100.00
& 99.92 \pm 0.58
& 5.50 \pm 1.81
& 6.50 \pm 10.05
& \bfseries 98.67
& \bfseries 99.36 \pm 3.01
& 5.72 \pm 1.80
& 6.78 \pm 10.56 \\

\bottomrule
\end{tabular}
\end{table}

The separation emerges in Route, where the operator can continue along the indicated path but the robot--payload system must depart from it. Across
both training and held-out maps, CARE exhibits the strongest task continuity in this condition, with the highest observed success and service
availability. DWA is the closest comparison method on these outcomes, but it exhibits substantially higher jerk-proxy values and substantially greater measured controller latency under the matched evaluation protocol. The contrast between the near-saturated Side results and the separated Route results shows that CARE becomes consequential when persistent operator guidance and executable robot motion temporarily diverge.

\subsection{Distinct Signatures of Reciprocal Refinement and Progress Recovery}
\label{sec:regime_ablation}

Table~\ref{tab:ablation} isolates the roles of CARE's two control
mechanisms by factorially enabling reciprocal refinement and progress
recovery around the same frozen nominal policy. The two conditions expose
different controller demands. Side preserves a locally usable
operator-indicated direction and therefore tests regulation within the
bounded reciprocal neighborhood. Route removes that local agreement and
tests whether the controller can leave a low-mobility configuration after
bounded refinement is no longer sufficient.

In Side, success is saturated for every configuration, and enabling recovery alone changes comfort violation and operator wait only slightly. Reciprocal refinement, in contrast, reduces both interaction-quality measures with and without recovery. In Route, recovery produces the dominant change in stall-sensitive outcomes, improving success and sharply reducing operator wait under both reciprocal settings, whereas reciprocal refinement consistently reduces comfort violation. The cross-condition pattern separates the two mechanisms empirically. Reciprocal refinement regulates interaction around the current base proposal, whereas progress recovery replaces the PPO base when the local neighborhood becomes persistently unproductive.

\begin{table}[ht]
\centering
\caption{Factorial ablation of reciprocal refinement and progress recovery
on held-out Side and Route maps, with 300 seed-matched episodes per
configuration--condition cell. Best results within each condition are
shown in bold.}
\label{tab:ablation}

\fontsize{7.0}{7.8}\selectfont
\setlength{\tabcolsep}{2.5pt}
\renewcommand{\arraystretch}{1.10}

\begin{tabular}{cc ccc ccc}
\toprule
\multicolumn{2}{c}{\textbf{Modules}}
& \multicolumn{3}{c}{\textbf{Side}}
& \multicolumn{3}{c}{\textbf{Route}} \\
\cmidrule(lr){1-2}
\cmidrule(lr){3-5}
\cmidrule(lr){6-8}

\textbf{Recip.}
& \textbf{Recov.}
& \textbf{Succ.} $\uparrow$
& \textbf{Comfort viol.} $\downarrow$
& \textbf{Wait} $\downarrow$
& \textbf{Succ.} $\uparrow$
& \textbf{Comfort viol.} $\downarrow$
& \textbf{Wait} $\downarrow$ \\

&
& \textbf{(\%)}
& \textbf{(\%)}
& \textbf{(s)}
& \textbf{(\%)}
& \textbf{(\%)}
& \textbf{(s)} \\
\midrule

\redxmark
& \redxmark
& \textbf{100.00}
& 6.89 $\pm$ 6.96
& 0.32 $\pm$ 1.09
& 96.33
& 5.99 $\pm$ 6.58
& 1.71 $\pm$ 11.46 \\

\greencmark
& \redxmark
& \textbf{100.00}
& 5.18 $\pm$ 5.36
& 0.24 $\pm$ 0.77
& 97.67
& 5.02 $\pm$ 5.46
& 1.63 $\pm$ 11.57 \\

\redxmark
& \greencmark
& \textbf{100.00}
& 6.91 $\pm$ 6.94
& 0.27 $\pm$ 1.04
& 98.33
& 6.11 $\pm$ 6.60
& 0.41 $\pm$ 1.24 \\

\greencmark
& \greencmark
& \textbf{100.00}
& \textbf{4.79 $\pm$ 5.34}
& \textbf{0.21 $\pm$ 0.61}
& \textbf{98.67}
& \textbf{4.53 $\pm$ 5.46}
& \textbf{0.38 $\pm$ 1.69} \\

\bottomrule
\end{tabular}
\end{table}

\subsection{Effect of Executable-Action Conditioning}
\label{sec:conditioning_ablation}

We isolate the response-conditioning choice while keeping all remaining
CARE components fixed. Shared predicts the operator response once under
the realized base action and reuses that prediction across candidates.
Raw-conditioned (pre-realization) conditions each candidate response on
its corresponding pre-realization proposal \(\mathbf v\), whereas
Executable-action conditions on the corresponding realized command
\(\mathcal R_t(\mathbf v;\mathbf o_t)\). The latter is the conditioning
strategy used by the full CARE controller.

All variants use the same frozen response checkpoint, nominal policy,
candidate-generation procedure, realization map, interaction utility,
acceptance rule, and recovery logic. The response predictor was trained
using post-realization executed actions as conditioning inputs.
Accordingly, Raw-conditioned is treated as a deliberate pre-realization
conditioning stress test rather than as a pure causal ablation of
realization consistency: differences from Executable-action conditioning
may reflect both realization inconsistency and covariate shift in the
predictor input distribution. We therefore do not interpret this
comparison as an isolated causal estimate of realization consistency.

Each variant is evaluated as an independent closed-loop controller on the
same seed-matched held-out Route episodes. Actions selected by one variant
are not reused by another.

Reciprocal activation denotes the fraction of control steps at which
bounded reciprocal search is activated. Realization exposure is the
fraction of activated steps for which realization modifies at least one
candidate. Selected realization gap is the episode mean of
\(\|\mathbf v_t^{\mathrm{sel}}-\mathbf u_t^{\mathrm{sel}}\|_2\)
over activated steps, where
\(\mathbf u_t^{\mathrm{sel}}
=\mathcal R_t(\mathbf v_t^{\mathrm{sel}};\mathbf o_t)\).

\begin{figure}[ht]
    \centering
    \includegraphics[width=\textwidth]
    {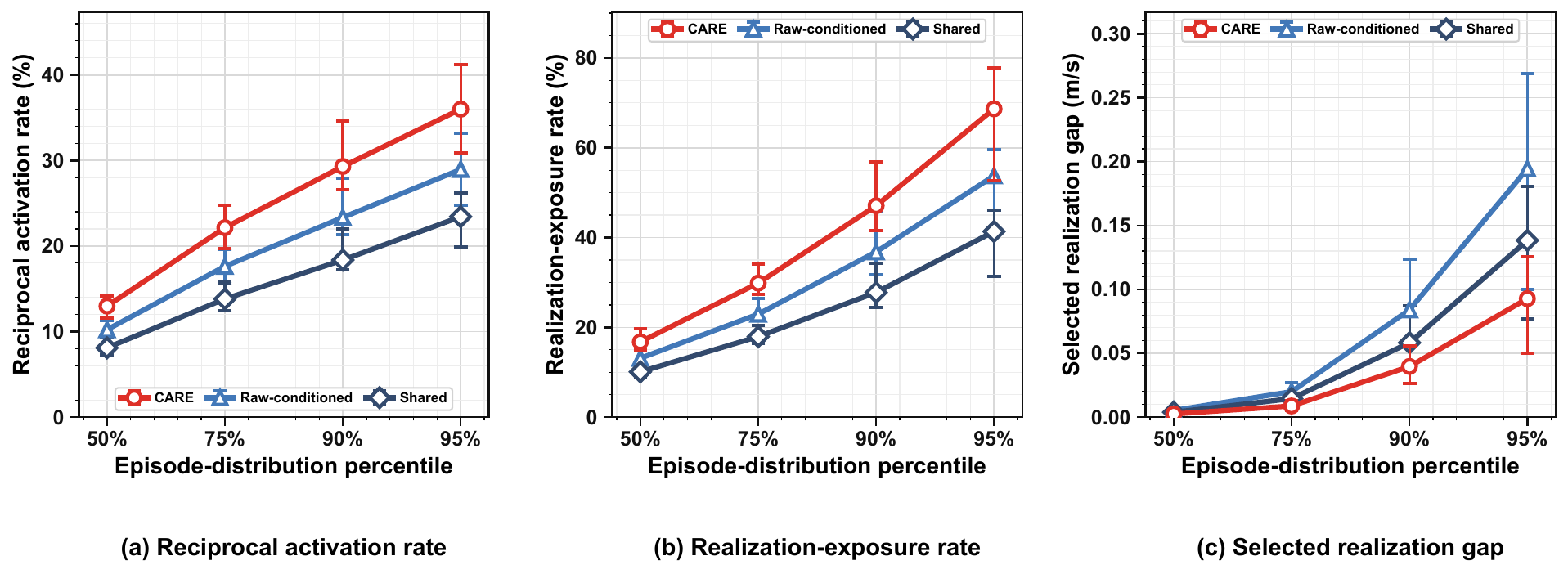}
    \caption{Effect of response conditioning on reciprocal activation,
    realization exposure, and selected realization gap over held-out Route
    episodes. Error bars show bootstrap 95\% confidence intervals}
    \label{fig:action_response_mechanism}
\end{figure}

Figure~\ref{fig:action_response_mechanism} characterizes how response
conditioning changes the deployed reciprocal process. Reciprocal
activation and realization exposure remain relatively close near the
median but separate progressively in the upper tail, with Shared showing
the lowest levels, Raw-conditioned intermediate levels, and
Executable-action the highest. Selected realization gap exhibits a
different pattern: Raw-conditioned develops the largest upper-tail
discrepancy, consistent with candidate rankings based on pre-realization
proposals that are subsequently modified before execution.
Executable-action conditioning reduces this discrepancy while
maintaining greater reciprocal engagement. The conditioning choice
therefore has its largest observed effect in episodes with repeated
reciprocal activation and realization exposure rather than uniformly
across routine operation.

The mechanism diagnostics alone do not establish whether these changes
affect closed-loop outcomes. Table~\ref{tab:conditioning_performance}
therefore reports task-level performance from the independently executed
closed-loop variants. Because only response conditioning is changed, the
Executable-action variant is identical to the full CARE configuration
and reproduces the corresponding CARE results in
Tables~\ref{tab:main_results} and~\ref{tab:ablation}.
\begin{table}[ht]
\centering
\caption{Task-level effect of response conditioning on held-out Route
maps over 300 seed-matched episodes. Continuous metrics report
mean \(\pm\) standard deviation. Best results are shown in bold.}
\label{tab:conditioning_performance}

\fontsize{7.4}{8.1}\selectfont
\setlength{\tabcolsep}{4.2pt}
\renewcommand{\arraystretch}{1.12}

\begin{tabular}{@{}lccc@{}}
\toprule
\textbf{Response conditioning}
& \textbf{Succ.} $\uparrow$
& \textbf{Serv.} $\uparrow$
& \textbf{Wait} $\downarrow$ \\
& \textbf{(\%)}
& \textbf{(\%)}
& \textbf{(s)} \\
\midrule

Shared
& 97.67
& 97.72 $\pm$ 4.84
& 0.81 $\pm$ 2.31 \\

Raw-conditioned
& 98.33
& 98.87 $\pm$ 3.89
& 0.67 $\pm$ 1.98 \\

\rowcolor{black!8}
\textbf{Executable-action (CARE)}
& \textbf{98.67}
& \textbf{99.36 $\pm$ 3.01}
& \textbf{0.38 $\pm$ 1.69} \\

\bottomrule
\end{tabular}
\end{table}

Task success remains near saturation across all three variants, whereas
service availability and operator waiting exhibit a consistent monotonic
ordering from Shared to Raw-conditioned and then to Executable-action
conditioning. Together with Fig.~\ref{fig:action_response_mechanism},
this pattern is consistent with executable-action-aligned conditioning
supporting greater service continuity and reduced operator interruption.
Because Raw-conditioned also changes the predictor's conditioning-input
distribution relative to training, the comparison is interpreted as
evidence for the practical importance of executable-action conditioning
rather than as a pure causal decomposition of realization consistency.

\subsection{Information Supporting the Nominal Proposal}

\begin{figure}[ht]
    \centering
    \includegraphics[width=\columnwidth]{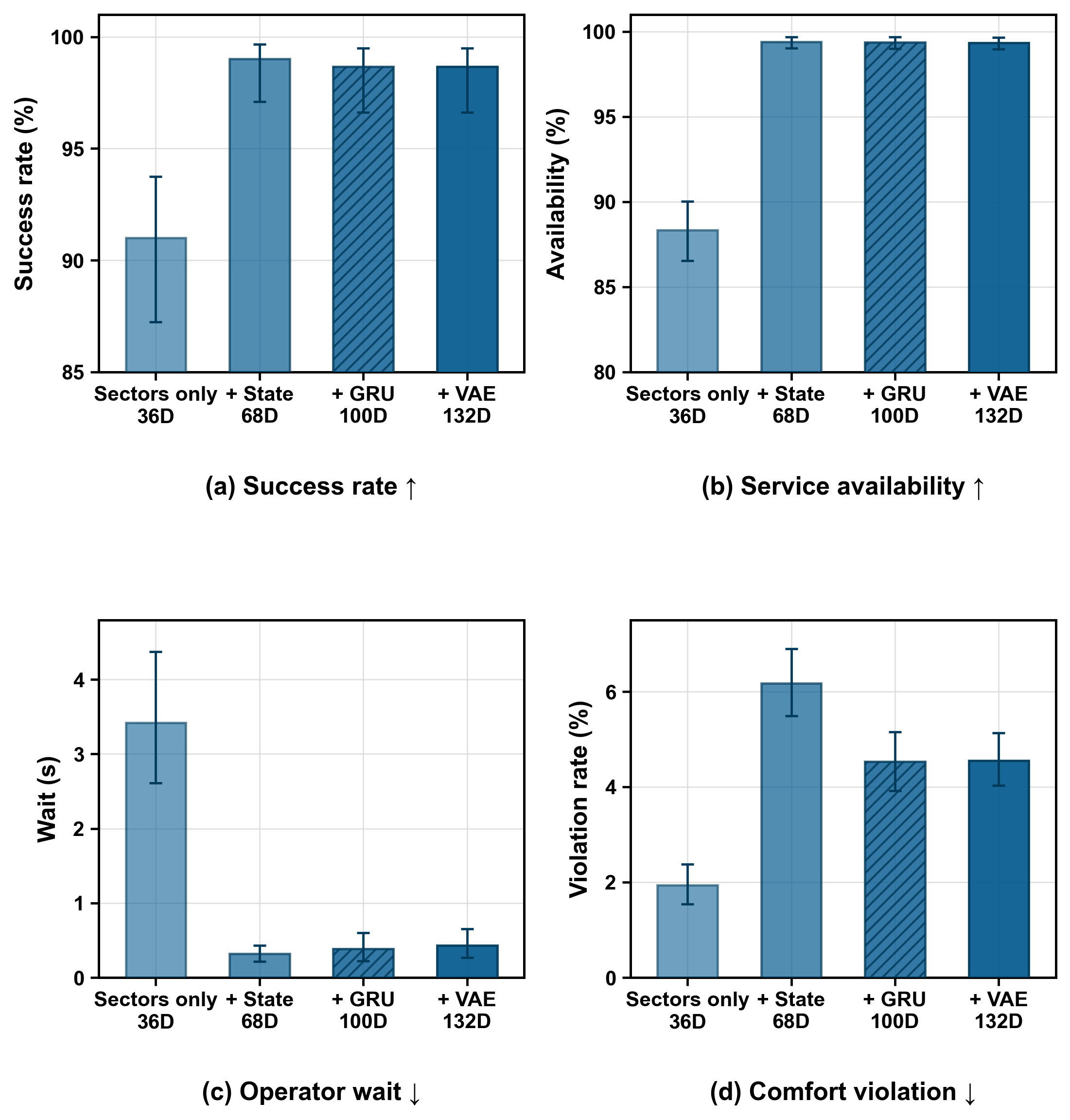}
    \caption{Cumulative observation-representation study for the nominal
    policy over pooled held-out Side and Route episodes. The four layouts
    are independently trained using the same training pipeline and evaluated
    under the same matched-seed protocol. Error bars show bootstrap 95\%
    confidence intervals}
    \label{fig:representation_ablation}
\end{figure}

CARE performs bounded refinement around a frozen nominal proposal.
Figure~\ref{fig:representation_ablation} examines the observation
representation used to produce that proposal. Each layout is trained independently using the same three-stage
pipeline---behavior cloning (BC), DAgger~\cite{ross2011reduction},
and PPO~\cite{schulman2017proximal}---and is evaluated under a common
matched-seed protocol. 

The study therefore measures the performance of separately retrained representations rather than sensitivity to input masking. The deployed nominal policy uses a 100-D representation comprising
36 direct LiDAR sectors, a 32-D encoding of the current robot--operator
state, and a 32-D GRU encoding of recent interaction history. The four layouts form a cumulative representation study: direct LiDAR sectors only (36-D), sectors with the current robot--operator state
(68-D), sectors with state and GRU history (100-D), and the same representation with an additional variational autoencoder (VAE) branch~\cite{kingma2014autoencoding} (132-D). The VAE branch is evaluated only as an additional representation candidate and is not used by the deployed policy.

The largest task-continuity change occurs when the current
robot--operator state is added to the direct LiDAR sectors: success and
service availability increase while operator wait decreases, indicating
that local traversability alone is insufficient for producing a useful
collaborative proposal. Adding the history-GRU branch preserves these
gains while reducing comfort violation, whereas adding the VAE branch
does not improve the separately retrained policy. Overall, the results
support the deployed 100-D representation combining direct local
geometry, current robot--operator state, and recent interaction history.

\subsection{Robustness to Delayed Operator Adaptation}
\label{sec:sensitivity}

For this study, the tracking-loss-or-stall rate is defined as the fraction
of episodes that either terminate because operator tracking is lost or
contain at least one command-stall event. The two events are combined by
logical union, so an episode satisfying both conditions is counted once.
Operator-response delay perturbs the temporal alignment between an executed
robot action and the operator adaptation induced by that action.

\begin{figure}[htbp]
    \centering
    \includegraphics[width=\columnwidth]{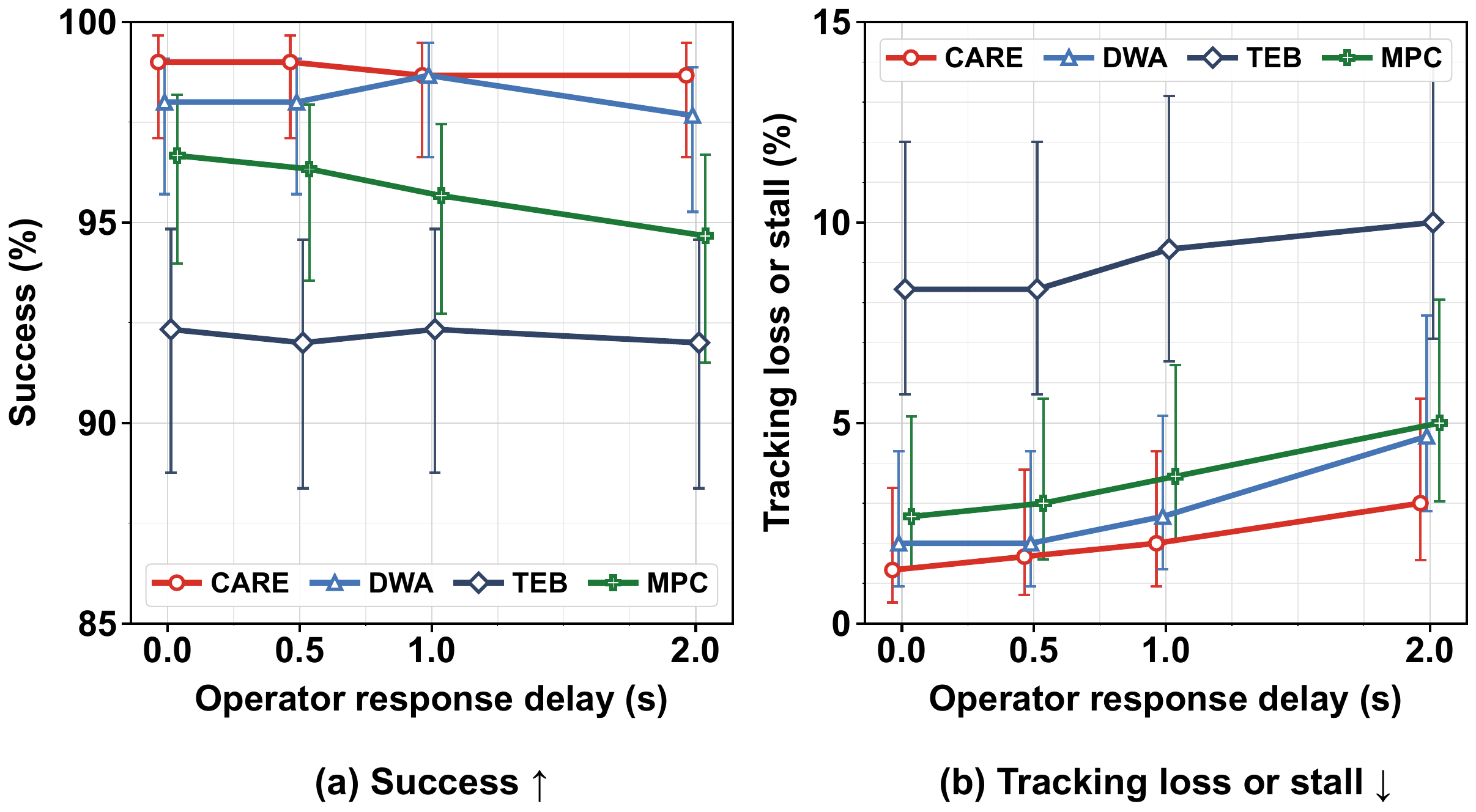}
    \caption{Robustness to operator-response delay on held-out Route maps.
    Error bars show two-sided Wilson 95\% confidence intervals}
    \label{fig:delay_sensitivity}
\end{figure}

Figure~\ref{fig:delay_sensitivity} evaluates this mismatch on held-out Route
episodes at delays of 0, 0.5, 1.0, and 2.0~s while keeping all learned
models and controller parameters fixed. The intervention therefore changes
only when the behavioral consequence of robot execution enters the closed
loop.

Across the tested delay range, terminal success changes less strongly than
the tracking-loss-or-stall rate. Temporal mismatch therefore appears first
as degraded interaction continuity rather than immediate mission failure.
CARE maintains high success while exhibiting the lowest observed
tracking-loss-or-stall rate throughout the tested range, indicating that
the complete closed loop remains effective under the evaluated
non-instantaneous responses.

\subsection{Robustness to Sensing-Range Variation}
\label{sec:lidar_sensitivity}

A supporting sensitivity study examines CARE's behavior under changes in
the LiDAR sensing horizon. On held-out Route maps, the maximum sensing
range is varied from 3 to 10~m while the learned models, controller
parameters, and remaining evaluation settings are fixed. The variation
changes the available local geometric context while preserving the
underlying route--feasibility mismatch.

\begin{figure}[htbp]
    \centering
    \includegraphics[width=\columnwidth]{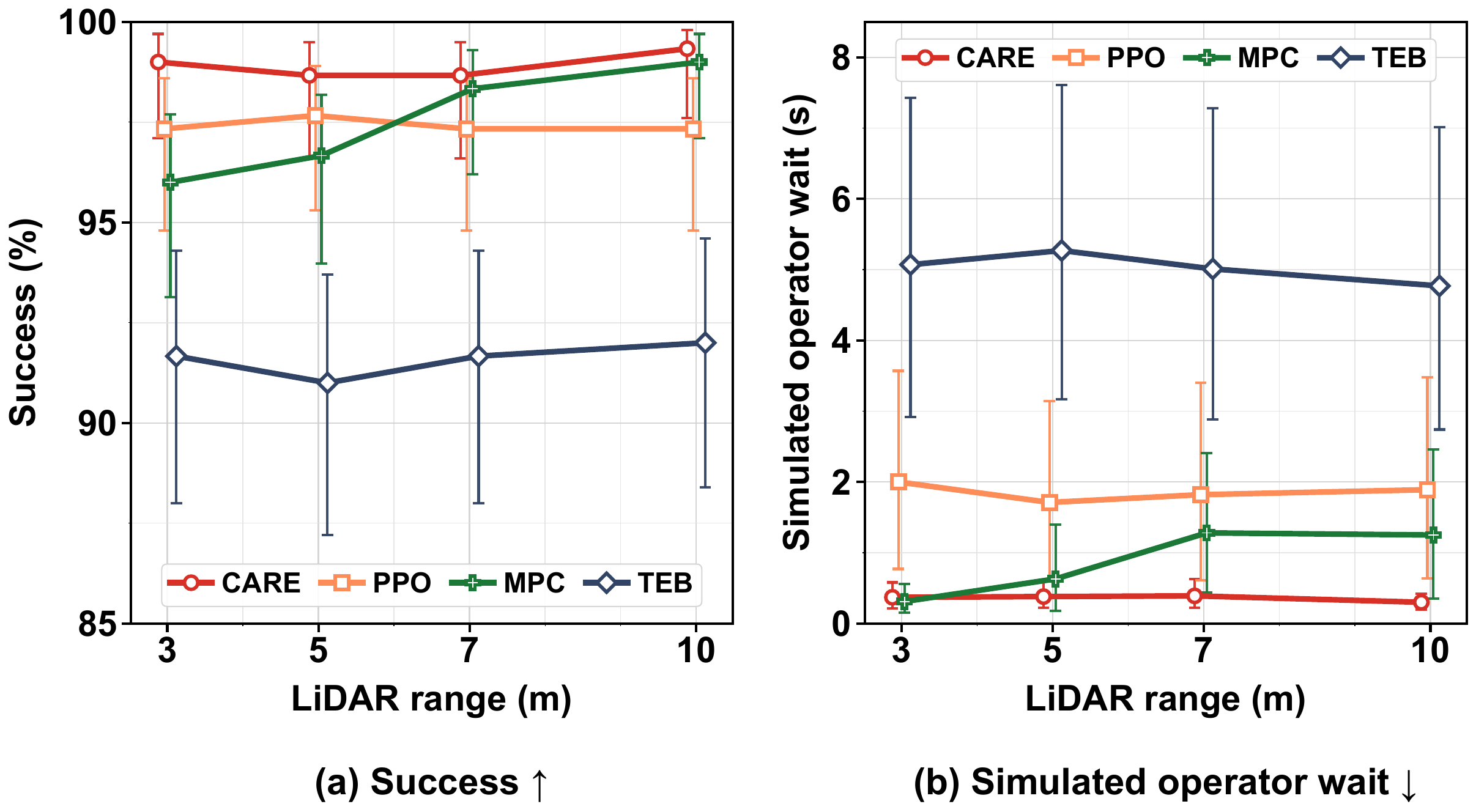}
    \caption{Robustness to LiDAR sensing range on held-out Route maps.
    Panels report task success and operator wait. Error bars show
    two-sided Wilson 95\% confidence intervals for success and
    episode-level percentile-bootstrap 95\% confidence intervals for
    operator wait}
    \label{fig:lidar_sensitivity}
\end{figure}

Figure~\ref{fig:lidar_sensitivity} shows that CARE preserves a stable
success--waiting profile across the tested sensing ranges. Its task
success remains near saturation, while operator wait remains consistently
low. The PPO reference is similarly stable in success but requires more
operator waiting. MPC benefits more visibly from additional geometric
context in terms of task completion, although this improvement is
accompanied by increased operator wait. TEB remains separated from CARE
on both outcomes. CARE's behavior is therefore not tied to a single
sensing-range setting, and its favorable balance between task completion
and operator continuity persists throughout the evaluated interval.

\subsection{Proof-of-Concept Physical Execution}

The physical execution shown in Fig.~\ref{fig:realworld} provides a
proof-of-concept demonstration of the CARE loop on a Clearpath Ridgeback
omnidirectional mobile base, with a Franka robotic arm mounted on the
platform but not involved in the control task. The deployed system uses
the frozen nominal policy and combines onboard operator tracking, local
LiDAR perception, online action realization, and planar velocity
execution. In the illustrated execution, CARE maintains human-guided
following while the robot locally departs from the operator-indicated
motion to negotiate the visible obstacle. The resulting trajectory
demonstrates that the complete decision loop can execute online under
physical sensing and actuation. This execution is intended as a
feasibility demonstration rather than a quantitative physical benchmark.

\FloatBarrier
\section{Conclusion}

Human-guided robot transport becomes a reciprocal decision problem when the
operator's route intention and the robot's executable motion temporarily
diverge. In this setting, predicting the operator under one robot action while
evaluating or executing another breaks the interaction being modeled. We
formulate the resulting requirement as action--response consistency: each
candidate must be paired with its own expected operator response, while
response prediction, joint rollout, interaction evaluation, and execution
must refer to the same executable action. CARE implements this principle by
evaluating executable-action-indexed joint human--robot transitions after
action realization. Bounded reciprocal refinement regulates transient
route--feasibility mismatch around the current base proposal, whereas
persistent low mobility temporarily replaces the nominal PPO proposal with a
progress-oriented DWA proposal and re-centers the same reciprocal evaluation
process.

The experiments exhibit the interaction-selective pattern predicted by this
formulation. CARE preserves near-saturated behavior when operator guidance
remains locally usable and achieves the highest observed task success and
service availability when the operator can continue along a route that the
robot--payload system must leave. The factorial ablation shows complementary
roles for reciprocal refinement and progress recovery, while the
executable-action-conditioning study demonstrates that the conditioning
choice changes the deployed reciprocal process most clearly in the upper tail
of the episode distribution. Delayed-response and sensing-range studies
further support closed-loop robustness under the evaluated perturbations, and
the proof-of-concept deployment demonstrates physical executability. Together, these results support action--response-consistent execution as a practical basis for reciprocal human-guided robot transport. Future work will extend the formulation to richer planar operator-response models, payload dynamics, and repeated multi-operator physical evaluation.

\section*{Statements and Declarations}

\noindent\textbf{Funding.}
This research did not receive funding.

\medskip
\noindent\textbf{Competing Interests.}
The authors have no relevant financial or non-financial interests to disclose.

\medskip
\noindent\textbf{Data Availability.}
The data supporting the findings of this study are available from the
corresponding author upon reasonable request. The implementation, trained
checkpoints, and frozen evaluation configurations will be made publicly
available upon publication.

\medskip
\noindent\textbf{Author Contributions.}
N.L.: Conceptualization, methodology, software, investigation,
experiments, analysis, and original manuscript preparation.
S.S.: Writing suggestions and figure design refinement.
Z.L.: Figure preparation and manuscript assistance.
M.D.: Technical guidance, research discussion, and manuscript feedback.
J.J.C.: Resources, evaluation metric guidance, supervision, and manuscript
review and editing.
L.W.: Technical guidance and manuscript feedback.
T.B.: Supervision, resources, and manuscript review and editing.

\medskip
\noindent\textbf{Consent to Publish.}
The individual shown in Fig.~1 provided consent for publication of the image.
\bibliographystyle{unsrtnat}
\bibliography{bibliography/ref}

\end{document}